\documentclass[runningheads]{llncs}
\usepackage{cite}
\usepackage{adjustbox}
\usepackage{array}
\usepackage{pifont}
\usepackage{xcolor} 
\usepackage{ulem} 
\usepackage{multirow}
\usepackage{booktabs}
\usepackage{amsmath}
\usepackage{graphicx}
\usepackage{CJKutf8} 
\usepackage{cleveref} 
\usepackage{algorithm}
\usepackage{algpseudocode}
\begin{document}
\title{Ensemble Genetic Programming}
%
%
\author{Nuno M. Rodrigues\inst{}\orcidID{0000-0001-5312-8276} \and
Jo\~ao E. Batista\inst{}\orcidID{0000-0002-2997-8550} \and
Sara Silva\inst{}\orcidID{ 0000-0001-8223-4799}}

\authorrunning{Rodrigues, N., Batista, J., Silva, S.}

\institute{Faculdade de Ci\^encias da Universidade de Lisboa, Lisboa, Portugal 
5\email{nmrodrigues@fc.ul.pt}\\
\and
Faculdade de Ci\^encias da Universidade de Lisboa, Lisboa, Portugal 
\email{jebatista@fc.ul.pt}\\
\and
Faculdade de Ci\^encias da Universidade de Lisboa, Lisboa, Portugal\\ 
\email{sara@fc.ul.pt}}

\institute{
Faculdade de Ci\^encias da Universidade de Lisboa, Lisboa, Portugal 
\email{\{nmrodrigues,jebatista,sara\}@fc.ul.pt}\\
}

\maketitle              
\begin{abstract}
Ensemble learning is a powerful paradigm that has been used in the top state-of-the-art machine learning methods like Random Forests and XGBoost. Inspired by the success of such methods, we have developed a new Genetic Programming method called Ensemble GP. The evolutionary cycle of Ensemble GP follows the same steps as other Genetic Programming systems, but with differences in the population structure, fitness evaluation and genetic operators. We have tested this method on eight binary classification problems, achieving results significantly better than standard GP, with much smaller models. Although other methods like M3GP and XGBoost were the best overall, Ensemble GP was able to achieve exceptionally good generalization results on a particularly hard problem where none of the other methods was able to succeed.

\end{abstract}

\section{Introduction}

Genetic Programming (GP)~\cite{Poli:2008:FGG:1796422} is one of the most proficient Machine Learning (ML) methods. It is capable of addressing multiple tasks such as classification and regression, using a variety of techniques from the most classical~\cite{Koza1992} to the most recent, like the geometric semantic approaches~\cite{Vanneschi} and the cluster-based multiclass classification~\cite{Munoz}.

Ensemble learning~\cite{Thomas} is a powerful ML paradigm where multiple models are induced and their predicted outputs are combined in order to obtain predictions that are more accurate than the individual ones. Some of the most successful ML methods are based on ensemble learning, like Random Forests (RF)~\cite{Breiman2001} and XGBoost (XG)~\cite{2016XGBoostAS}. On the other hand, their performance may vary substantially depending on the setting of some crucial parameters, like the number of trees and their maximum depth, which in turn depend on the properties of each dataset.

Inspired by the success of such methods, and motivated by the need to automatically find the right settings for these parameters, we have developed a new GP method called Ensemble GP (eGP). The evolutionary cycle of eGP follows the same steps as other GP systems, but with differences in the population structure, fitness evaluation and genetic operators. In particular, the population is composed of two subpopulations, trees and forests, where each subpopulation uses its own fitness function and genetic operators. The approach can be described as co-evolutionary, cooperative and compositional, and involves subsampling of both observations and features.

The rest of the paper is organized as follows. Section~\ref{section:related_work} describes related work, while Section~\ref{section:eGP} provides the details of the eGP method. Section~\ref{section:experimental_setup} specifies the experimental setup, and Sections~\ref{section:results} and~\ref{section:discussion} report and discuss the results obtained. Finally, Section~\ref{section:conclusions} contains the conclusions and future work.

\section{Related Work}
\label{section:related_work}

Evolutionary computation and other bio-inspired methods have been linked to ensemble learning from early on (see~\cite{Gagne:2007:ELF:1276958.1277317} and references therein). An obvious way to build ensembles is to combine different individuals of a population, whether they are GP individuals (\textit{e.g.}~\cite{Zhang1996EnhancingRO}) or other types, like neural networks (review in~\cite{Islam2008EvolvingAN}). Many other types of ensembles have been built using evolutionary and other bio-inspired methods, like ensembles of clustering algorithms~\cite{Coelho:2011:MDH:2304780.2304934}, Decision Trees~\cite{Cantu}, Support Vector Machines~\cite{PADILHA201685}, or a mix of different types~\cite{Escalante}. Diversity is important among ensemble members, and multiobjective evolutionary approaches have been often used to address this issue (\textit{e.g.}~\cite{Oliveira2009UseOM,6198882,Chandra2006} and references therein). 

A multitude of publications focus on single specific aspects of ensemble learning, like selecting and combining the members of the ensemble (\textit{e.g.}~\cite{Escalante} and references therein), or evolving the functions that combine the different members (\textit{e.g.}~\cite{Escalante,Langdon01geneticprogramming,1716247}). Others focus on building complete ensembles from scratch, but even if we limit ourselves to the ones that use GP exclusively (\textit{e.g.}~\cite{Iba:1999:BBB:2934046.2934063,Brameier:2001:ETP:594855.594901, Veeramachaneni2015}), we find a large diversity of designs, goals and scales of application. A systematic review of this extremely vast and diverse literature is much needed in both evolutionary and ensemble learning communities.

\section{Ensemble GP}
\label{section:eGP}
Now, we describe the method we call ensemble GP (eGP) with all the variants we implemented and tested. The evolutionary cycle of eGP follows the same steps as other GP systems, but with differences in the population structure, fitness evaluation and genetic operators. In particular, the population is composed of two subpopulations, where each subpopulation uses its own fitness function and genetic operators. The approach can be described as co-evolutionary, cooperative and compositional, and involves subsampling of both observations and features. Algorithm~\ref{alg:ALG1} describes the main steps of eGP.

\begin{algorithm}
  \caption{eGP} \label{alg:ALG1}
  \begin{algorithmic}
  \Procedure{eGP} {$Dataset (D_{s}), n_{t}, n_{f}$}
    \State $Split\; D_{s}\; into\; training,\; testing\; and\; sub\,samples\; \Phi$
    \State  $T_{list} \gets Generate\;Trees(\Phi, n_{t})$
    \State  $F_{list} \gets Generate\;Forests(n_{f})$
    \While  {$generation (g)  < max\;generations$}
      \State  $T_{parents} \gets Selection(T_{list})$
      \State  $F_{parents} \gets Selection(F_{list})$
      \State  $T_{offspring} \gets Breeding(T_{parents})$
      \State  $F_{offspring} \gets Breeding(F_{parents})$
      \State  $F_{list} \gets Prune(F_{offspring})$\Comment{Prune only the best forest}
      \State $g++$
    \EndWhile
  \EndProcedure
  \end{algorithmic}
\end{algorithm}

Before describing the details regarding the population, fitness and genetic operators of eGP, we briefly describe a GP system called M3GP~\cite{Munoz} (Multidimensional Multiclass GP with Multidimensional Populations), not only because it is one of the baselines in our experiments, but also because some elements of eGP are highly inspired in M3GP.

\subsection{M3GP}

In terms of representation of the solutions, the main difference between M3GP and standard tree-based GP is the number of trees that are part of the same individual. While a standard GP individual is a single tree, a M3GP individual may be composed of several trees, called dimensions. Originally developed for performing multiclass classification~\cite{Ingalalli,Munoz}, M3GP evolves each individual as a set of hyperfeatures, each one represented by a different tree/dimension. After remapping the input data into this new multidimensional feature space, it calculates the accuracy by forming clusters based on the data labels and classifying each observation as the class of the closest centroid according to the Mahalanobis distance. M3GP has also been used for evolving hyperfeatures for regression~\cite{Munoz2019} and for classification in other GP systems~\cite{lacava}.

Starting with only one tree/dimension per individual, M3GP uses standard subtree crossover and mutation between individuals, and three other operators designed for removing a tree/dimension from an individual, adding a randomly created tree/dimension to an individual, and swapping trees/dimensions between individuals. Additionally, a pruning operator is applied to the best individual of each generation, removing the trees/dimensions that do not improve its accuracy.

\subsection{eGP Population Structure}
\label{subsection:population}

The population is composed of two types of individuals: trees and forests. A tree is not the output model, but only a part of it. The output model is a forest, built as an ensemble of trees. Each tree may be part of many different forests, and some trees may be part of none.

Trees have the same structure as those used in standard GP, but instead of having access to all the observations and features of the training dataset, each individual only sees a subset of observations, and in many cases also a subset of features. Different variants of eGP use different sampling options: 1) 60\% of all observations, all features included; 2) between one and all observations, one to all features included, these numbers being randomly chosen before each sampling. In both options, the sampling is done uniformly without replacement, and repeated whenever a new subset of training data is required for allocating to a new tree.

Forests have the same structure as the M3GP individuals, with each dimension being a tree from the subpopulation of trees.

\subsection{eGP Fitness Functions}

The subpopulation of trees uses a standard fitness function based on the error between expected and predicted outputs, like the Root Mean Squared Error (RMSE). In classification problems, the class labels are interpreted as the numeric expected outputs. The fitness of each tree is calculated using only the subset of observations allowed for this tree. 

The subpopulation of forests uses a fitness function based on the accuracy obtained on all the observations of the training set. Each forest gathers, for each observation, a vote (on a class) from each of the trees that compose its ensemble. This vote is obtained by adopting the class label that is closer to the predicted output. The votes from the different trees of the ensemble can be combined by normal majority voting or by weighted voting.

In normal voting, for each observation the class that receives more votes wins, and ties are solved by randomly choosing one of the classes. In weighted voting (Algorithm~\ref{alg:ALG2}), for each observation a certainty value is calculated for each class prediction of each tree, based on the vector of predicted values by all the trees of the ensemble~\eqref{eq:certainty}. The sum of certainty values for each class
is then calculated, and divided by the sum of certainty values for both classes. The class with highest results is chosen as the prediction.

We chose to use L2 normalization \eqref{eq:norm} for consistency with the cosine similarity used for the eCrossover (described next), which also uses L2. Other normalization methods were considered. Min-Max was discarded due to its inability for dealing with outliers; Z-Score was discarded because the resulting array was not contained in the $]0,1[$ range.

\begin{equation}
    certainty = 1 - l_2(X), 
    X = \begin{pmatrix}
        x_1 \\
        x_2 \\
        \vdots \\
        x_n
    \end{pmatrix}
    \label{eq:certainty}
\end{equation}

\begin{equation}
    l_2\;normalization = \sqrt{\sum_{k=1}^{n} |x_k|^2} 
    \label{eq:norm}
\end{equation}

\begin{algorithm}
  \caption{Weighted Voting} \label{alg:ALG2}
  \begin{algorithmic}
    \Procedure{weighted\_voting}{$predictions,certainties$}
      \State $votes\gets [\:]$
      \For{\texttt{$row$ $in$ $predictions$}}
        \State $zeros, ones\gets 0$
        \For{\texttt{$col$ $in$ $certainties$}}
            \State \textbf{if} $predictions[row][col] == 1$
                \State \quad $ones += certainties[row][col]$
            \State \textbf{else}
                \State \quad $zeros += certainties[row][col]$
        \EndFor
      \State $votes.append(0\: \textbf{if}\, zeros/(zeros+ones) \geq ones/(zeros+ones)\, \textbf{else}\: 1)$
      \EndFor
    \EndProcedure
  \end{algorithmic}
\end{algorithm}

\subsection{eGP Genetic Operators}

The trees and forests of eGP use different genetic operators. Trees use what can be described as protected versions of the standard subtree crossover and mutation, designated here as eCrossover and eMutation, respectively. The protection is needed when parent trees are not allowed to see all the features due to feature sampling (see Section~\ref{subsection:population}). In this case, the offspring must inherit feature restrictions from their parents, otherwise after a number of generations all the trees will be using all the features. Without feature sampling, these operators behave the same as the standard ones.

eMutation simply has to ensure that the new subtree created to replace a random branch of the parent is restricted to the same subset of features as the parent. eCrossover must guarantee that each swapped branch is also restricted to the subset of features inherited by the receiving offspring. Each of the two offspring inherits from one of its two parents. Instead of relying on a careful choice of compatible couples and branches to swap, eCrossover relies on a repair procedure that replaces features on the received branches whenever these features are not allowed by the inherited restrictions (Algorithm~\ref{alg:ALG3}). Each illegal feature is compared to all the legal ones on the complete training set, using the cosine similarity measure \eqref{eq:cosine}. The chosen replacement is the most similar feature to the one that was removed. Unlike the euclidean distance, the cosine similarity can compare and recognize two vectors of similar meaning even if they have very different magnitudes. 

%

\begin{equation}
    S(X,Y) = \frac{\sum_{k=1}^{n} |x_k||y_k|}{\sqrt{\sum_{k=1}^{n} |x_k|^2} \sqrt{\sum_{k=1}^{n} |y_k|^2}}
    \label{eq:cosine}
\end{equation}

\begin{algorithm}
  \caption{eCrossover}\label{alg:ALG3}
  \begin{algorithmic}
  \Procedure{subtree crossover}{$parent_{1}, parent_{2}$}
    \State  $cp_{1}, cp_{2} \gets choose\;crossover\;points$ \Comment{crossover point 1 and 2}
    \State  $refact\;tree(parent_{1}, parent_{2}, cp_{1}, cp_{2}, bag_{1}, bag_{2}, Data_{Training})$
  \EndProcedure
  \\
  \Procedure {refact tree}{$parent_{1}, parent_{2}, cp_{1}, cp_{2}, bag_{1}, bag_{2}, Data_{Training}$}
    \State  $parent_{1}, parent_{2} \gets swap\; branches(cp_{1}, cp_{2})$
    \State  $fix\; terminals(p_{1}, bag_{1}, bag_{2},Data_{Training})$     
    \State  $fix\;terminals(p_{2}, bag_{2}, bag_{1},Data_{Training})$
  \EndProcedure
  \end{algorithmic}
\end{algorithm}

Regarding the subpopulation of forests, it uses the same genetic operators as M3GP, namely two mutation operators to add and remove trees from the ensemble, and one crossover operator to swap trees between different ensembles.

\section{Experimental Setup}
\label{section:experimental_setup}
This section describes our experimental setup for the eGP methods, comparing them against two baselines, standard GP and M3GP, and two state-of-the-art classifiers, Random Forests (RF) and XGBoost (XG). Six different variants of eGP were tested, and the results were analysed in terms of training and test accuracy, number of trees and number of nodes of the final solutions. When comparing accuracy, statistical significance is determined using the non-parametric Kruskal-Wallis test at $p<0.01$. Next, we describe all the 10 methods tested, their main parameter settings, and the eight datasets used for obtaining the reported results.

\subsection{Methods}

Table \ref{table:methods} contains the acronyms and descriptions of all the methods used, and will serve as a memory aid for the remainder of this paper. The six eGP variants are eGP-N and eGP-W (normal and weighted voting with sampling of features and observations); eGP-N5 and eGP-W5 (same as previous but with populations of 500 trees and 500 forests, instead of 250 each); eGPn and eGPw (same as eGP-N and eGP-W but without feature sampling).

\begin{table}
\centering
\caption{Acronyms and descriptions of the methods}
\begin{tabular}{llllllll}
\hline
~GP       & \multicolumn{6}{l}{\multirow{11}{*}{}} & Standard Genetic Programming\\
~M3GP     & \multicolumn{6}{l}{}                   & Multidimensional Multiclass GP with Multidimensional Populations\\
\hline
~eGP-N    & \multicolumn{6}{l}{}                   & Ensemble GP, feature sampling, normal voting\\
~eGP-W    & \multicolumn{6}{l}{}                   & Ensemble GP, feature sampling, weighted voting\\
~eGP-N5   & \multicolumn{6}{l}{}                   & Ensemble GP, feature sampling, normal voting, larger population\\
~eGP-W5   & \multicolumn{6}{l}{}                   & Ensemble GP, feature sampling, weighted voting, larger population\\
~eGPn     & \multicolumn{6}{l}{}                   & Ensemble GP, no feature sampling, normal voting \\
~eGPw     & \multicolumn{6}{l}{}                   & Ensemble GP, no feature sampling, weighted voting \\
\hline
~RF       & \multicolumn{6}{l}{}                   & Random Forests \\                                                  
~XG       & \multicolumn{6}{l}{}                   & XGBoost \\
\hline
\end{tabular}
\label{table:methods}
\end{table}

\subsection{Parameters}
\label{subsection:parameters}
Table~\ref{table:parameters} summarizes the main parameters used in the GP-based methods and in the RF and XG methods. Each experiment is performed 30 times, with each run using a different partition of the dataset in 70\% training and \%30 test. The GP-based methods run for 100 generations. GP and M3GP use populations of 500 individuals, while eGP initializes each subpopulation with 250 (or 500) individuals, for a total of 500 (or 1000) trees $+$ forests. Trees are initialized using Ramped Half-and-Half, as suggested by Koza~\cite{Koza1992}, while forests are initialized in a similar fashion to M3GP, with only one tree per forest~\cite{Munoz}. The arithmetic operators of the function set are protected in the following way: when dividing a value by zero, we return the numerator; when trying to square root or logarithm a negative number, we return the number untouched. Therefore, the protection is to ignore the presence of the operator whenever it raises an exception. No constants are used. The fitness guiding the evolution is the RMSE in GP, and the accuracy in M3GP and eGP. In order to obtain the accuracy from GP, the predicted outputs are transformed into the closest numeric class labels. Selection for breeding is made with Double Tournament~\cite{Luke} in GP and M3GP, and regular tournament in eGP, size 5. Regarding genetic operators, the crossover/mutation probabilities are 0.95/0.05 for GP, and 0.5/0.5 for both M3GP and eGP. This means choosing between crossover and mutation with equal probability, but for M3GP and eGP forests the specific type of crossover or mutation must then be chosen, also with equal probability. Elitism guarantees that the best parent is copied into the new population.

Regarding RF and XG (last three rows of the table), both were 10-fold cross-validated for number of estimators and maximum depth, and RF was also cross-validated for the impurity criterion.

\begin{table}[t]
\centering
  \caption{Main parameter settings}
  \begin{tabular}{l*{1}{l}r}
    \hline                                                                             
    Runs                        & 30\\
    Generations                 & 100\\
    Population Size             & GP/M3GP = 500, ~eGP = \{250+250, 500+500\}\\
    Function Set                & $\{+, -, \times, /, \log, \sqrt\}$ ~(protected)\\
    Fitness                     & GP = RMSE, ~M3GP/eGP = Accuracy\\
    Selection                   & Tournament size 5 ~(GP/M3GP = Double Tournament)\\
    Crossover/Mutation          & GP = 0.95/0.05, ~M3GP/eGP = 0.5/0.5\\
    \hline
    Number of Estimators~~~~~  & \{50, 100, 150, 200\}\\
    Maximum Depth               & \{2, 4, 6, 8\}\\
    Impurity Measure            & RF = \{Gini, Entropy\}\\
    \hline
  \end{tabular}
  \label{table:parameters}
\end{table}

\subsection{Datasets}

Table~\ref{table:datasets} describes the main characteristics of the datasets used in our experiments. We have selected eight problems from various domains, all being binary classification tasks, with a different number of features and observations.

\textbf{BCW}, \textbf{HEART}, \textbf{IONO}, \textbf{PARKS.} \textit{Breast Cancer Wisconsin}, \textit{Heart Disease}, \textit{Ionosphere} and \textit{Parkinsons} are datasets included in the UCI ML repository~\cite{Linchman}.
    
\textbf{BRAZIL.} \textit{Brazil} is a dataset for detecting burned areas in satellite imagery, containing the radiance values of a set of pixels from a Landsat 8 OLI image over Brazil, and corrected unburned/burned labels~\cite{SILVA2018323}.
    
\textbf{GAMETES.} \textit{GAMETES\_Epistasis\_2-Way\_1000atts\_0.4H\_EDM-1\_EDM-1\_1} is a simulated Genome-Wide Association Studies (GWAS) dataset generated using the GAMETES tool~\cite{lacava}, available in OpenML~\cite{gijsbers}.
    
\textbf{PPI.} \textit{GRID/HPRD-unbal-HS} is a dataset built from a Protein-Protein Interaction benchmark of the human species~\cite{10.1093/bioinformatics/btq483}, containing the Resnik\textsubscript{Max} semantic similarity measure between each pair of proteins on three different semantic aspects~\cite{rita}.  
    
\textbf{SONAR.} \textit{sonar.all-data} is a dataset for binary classification of sonar returns, available in Kaggle~\cite{zhang}.

\begin{table}[t]
\centering
\caption{Number of features, observations and negative/positive ratio on each dataset. }
\resizebox{\textwidth}{!}{%
\begin{tabular}{llclclclclclclclc}
Datasets      &  & BCW &  & BRAZIL &  & GAMETES &  & HEART &  & IONO &  & PARKS &  & PPI   &  & SONAR \\ \hline
Features      &  & 11  &  & 8      &  & 1000    &  & 13    &  & 33   &  & 23    &  & 3     &  & 61    \\
Observations  &  & 683  &  & 4872   &  & 1600    &  & 270   &  & 351  &  & 195   &  & 31320 &  & 208  \\
Neg/Pos Ratio &  & 35/65 &  & 42/58 &  & 50/50 &  & 45/55 &  & 65/35 &  & 75/25 &  & 52/48 &  & 46/54 \\
\end{tabular}%
}
\label{table:datasets}
\end{table}

\section{Results}
\label{section:results}
\begin{figure}
	\centering
	\includegraphics[width=\linewidth]{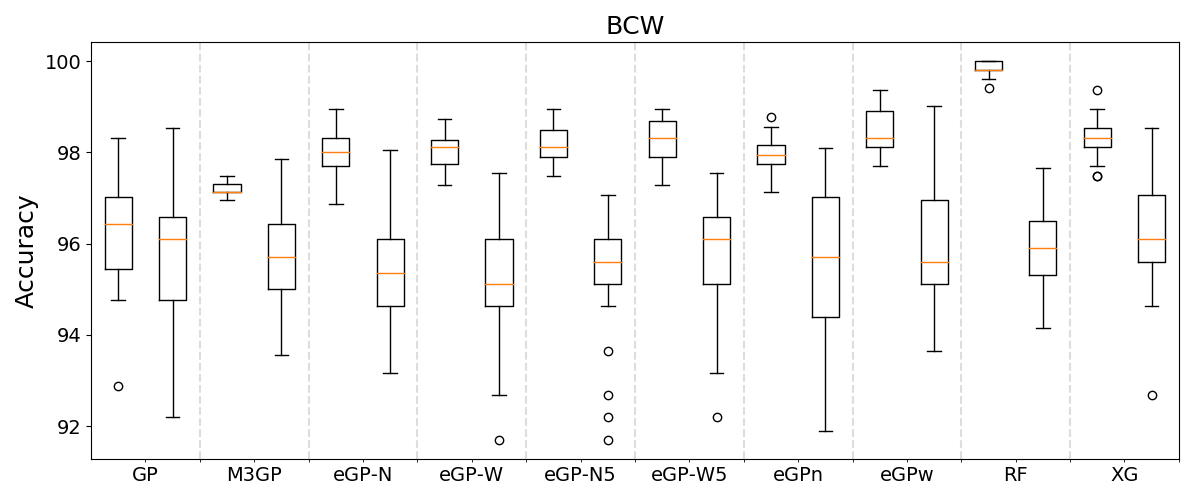}
	\vspace{-15pt}
	\caption{Boxplot for the training (left) and test (right) accuracy of each method in the BCW dataset.}
	\label{BCW_with_outliers}
\end{figure}

\begin{figure}
	\centering
	\includegraphics[width=\linewidth]{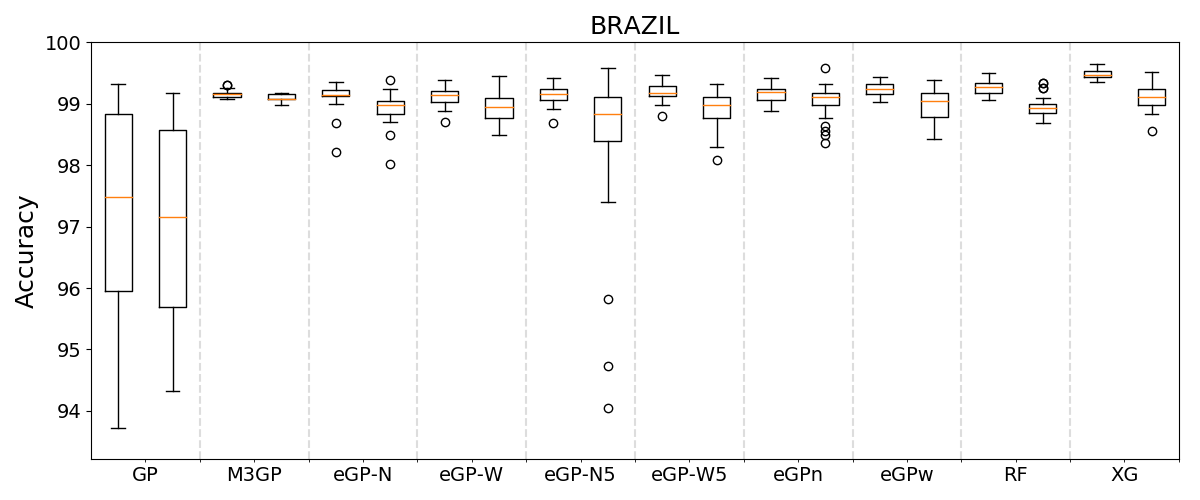}
	\vspace{-15pt}
	\caption{Boxplot for the training (left) and test (right) accuracy of each method in the BRAZIL dataset. Outliers removed for visualization purposes: on training, 90.97\% and 67.92\%, both for GP; on test, 90.70\% and 68.95\% for GP, and 77.29\% for eGP-N5.}
	\label{BRAZIL_with_outliers}
\end{figure}

\begin{figure}
	\centering
	\includegraphics[width=\linewidth]{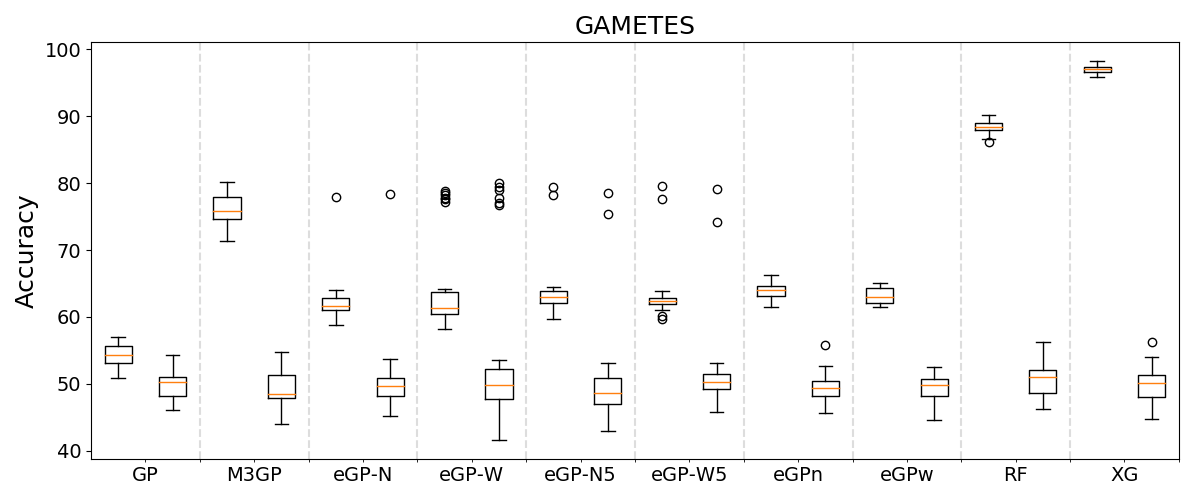}
	\vspace{-15pt}
	\caption{Boxplot for the training (left) and test (right) accuracy of each method in the GAMETES dataset.}
	\label{GAMETES_with_outliers}
\end{figure}

\begin{figure}
	\centering
	\includegraphics[width=\linewidth]{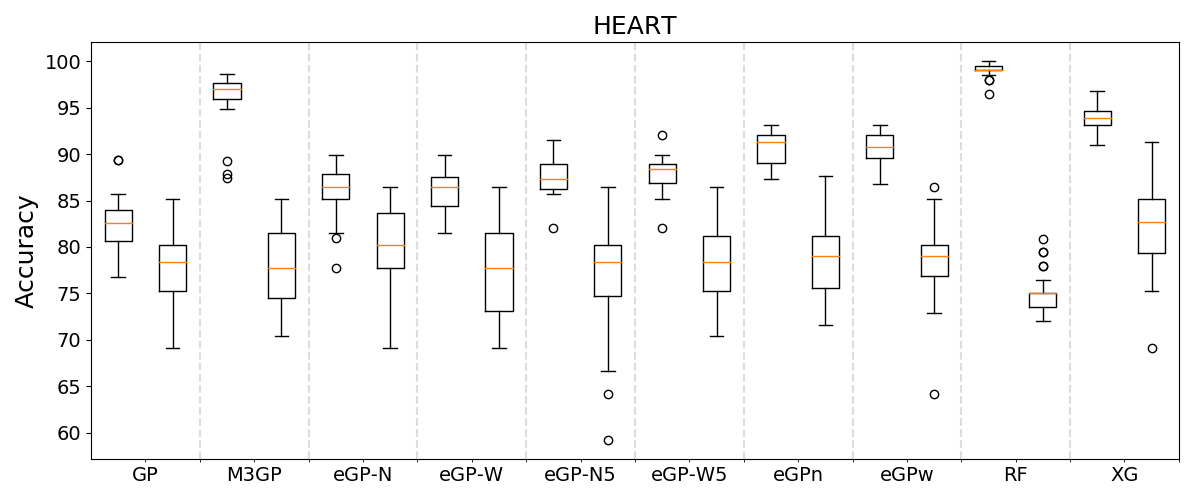}
	\caption{Boxplot for the training (left) and test (right) accuracy of each method in the HEART dataset.}
	\label{HEART_with_outliers}
\end{figure}

\begin{figure}
	\centering
	\includegraphics[width=\linewidth]{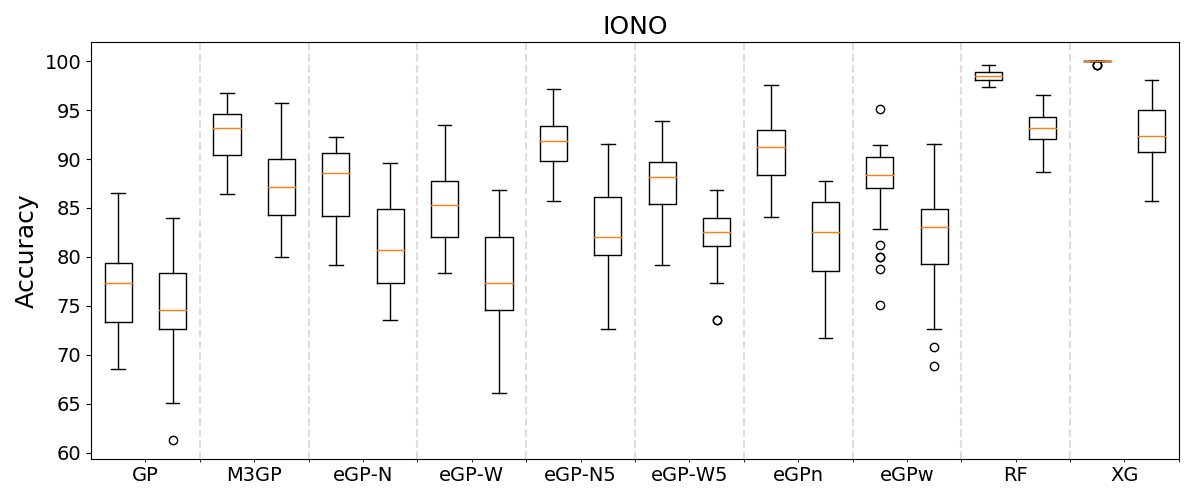}
	\caption{Boxplot for the training (left) and test (right) accuracy of each method in the IONO dataset.}
	\label{IONO_with_outliers}
\end{figure}

\begin{figure}
	\centering
	\includegraphics[width=\linewidth]{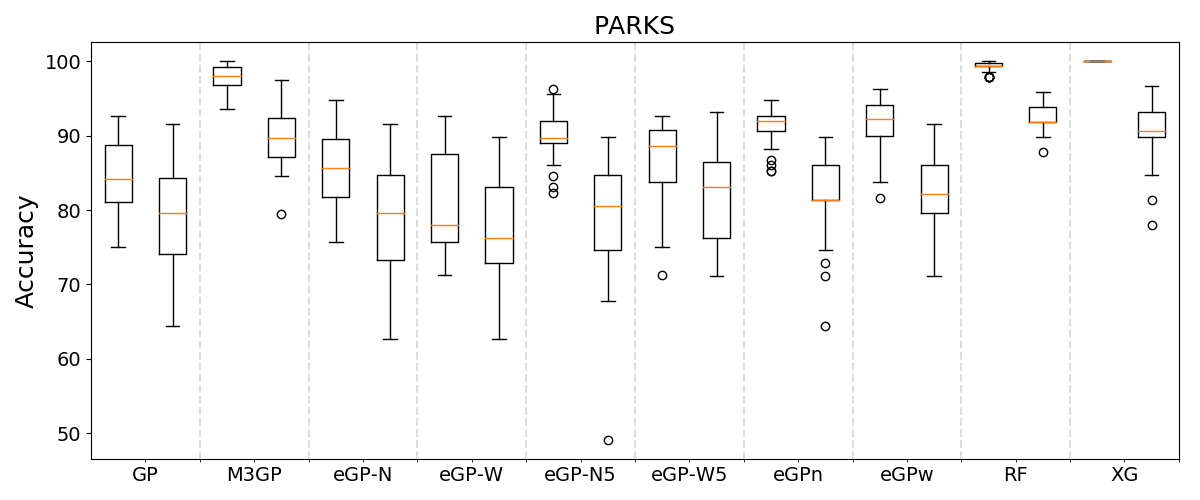}
	\caption{Boxplot for the training (left) and test (right) accuracy of each method in the PARKS dataset.}
	\label{PARK_with_outliers}
\end{figure}

\begin{figure}
\centering
	\includegraphics[width=\linewidth]{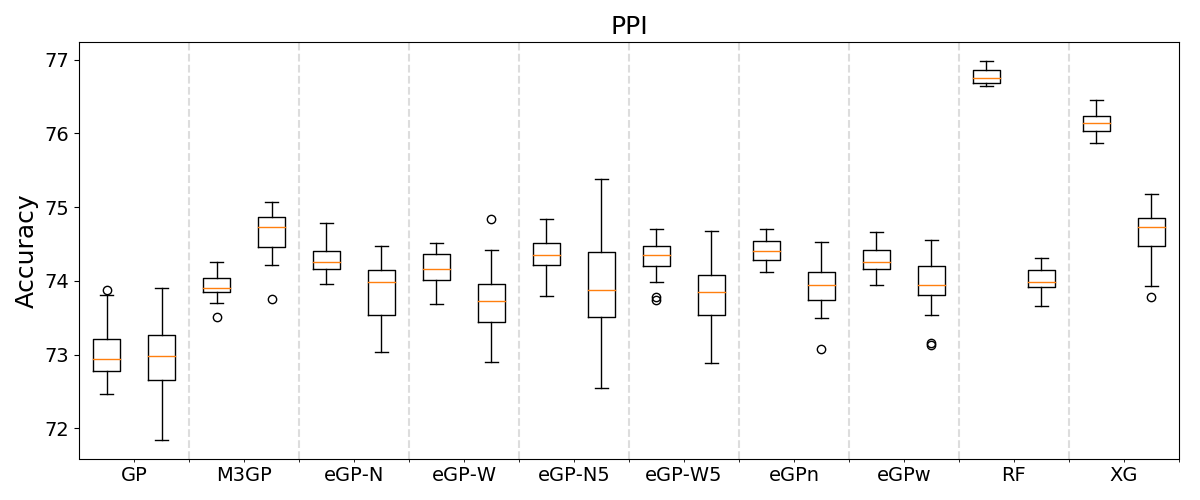}
	\vspace{-15pt}
	\caption{Boxplot for the training (left) and test (right) accuracy of each method in the PPI dataset.}
	\label{PPI_with_outliers}
\end{figure}

\begin{figure}
\centering
	\includegraphics[width=\linewidth]{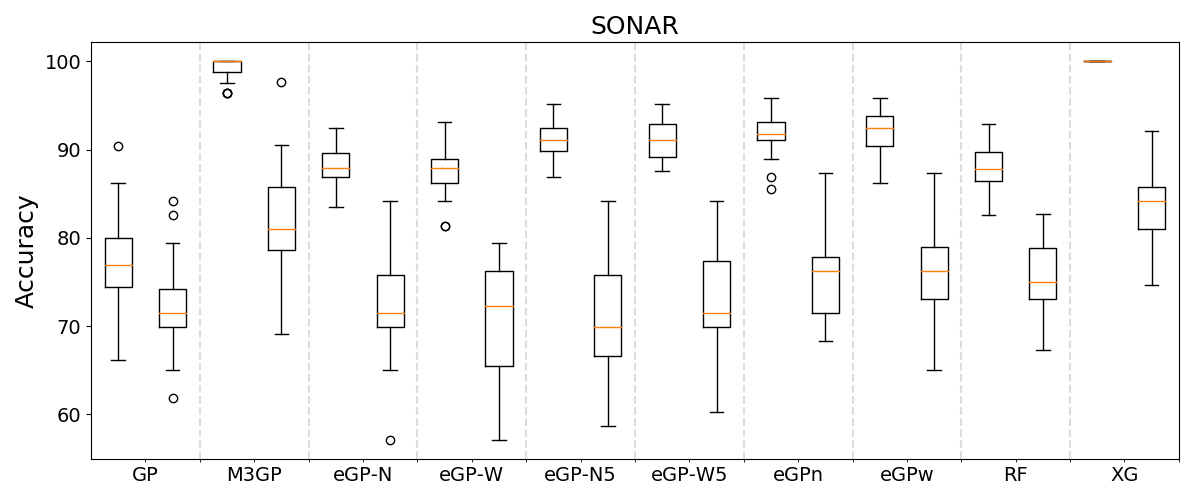}
	\vspace{-15pt}
	\caption{Boxplot for the training (left) and test (right) accuracy of each method in the SONAR dataset.}
	\label{SONAR_with_outliers}
\end{figure}

\Cref{BCW_with_outliers,BRAZIL_with_outliers,GAMETES_with_outliers,HEART_with_outliers,IONO_with_outliers,PARK_with_outliers,PPI_with_outliers,SONAR_with_outliers} show boxplots of the training and test accuracy obtained by all the methods on all the problems. For each problem there are two whiskered boxes, the left one for training and the right one for test. On the BRAZIL problem, five outliers were removed for visualization purposes, two on training (90.97\% and 67.92\%, both for GP) and three on test (90.70\% and 68.95\% for GP, 77.29\% for eGP-N5).

Between the two baselines, as expected M3GP is better than standard GP, achieving significantly better training accuracy on all eight problems, and also significantly better test accuracy on five of them (BRAZIL, IONO, PARKS, PPI and SONAR). In fact, in all pairwise comparisons with the other methods in all the problems, standard GP is significantly worse in 96\% of the cases on training, and 46\% on test. The only exception where it performs significantly better is on the HEART problem, against RF on the test data.

Regarding the two proposed methods eGP-N and eGP-W, a comparison between them reveals that the weighted voting (eGP-W) does not seem to improve performance over the normal voting (eGP-N), as the weighted voting resulted in one significantly worse training accuracy in the PARKS problem (and another borderline worse in IONO), all other results being equal to the ones of normal voting. Also between eGP-N5 and eGP-W5 the weighted voting resulted in one significantly worse training accuracy in the IONO problem, all other results showing no significant differences.

Increasing the population size from 250 to 500 proved to be only marginally beneficial, more to weighted than to normal voting.  eGP-N5 achieved significantly better results than eGP-N on four problems (GAMETES, IONO, PARKS and SONAR) on training, and none on test, all other results being statistically equal. eGP-W5 was significantly better than eGP-W on five problems (BCW, HEART, IONO, PARKS and SONAR) on training, and on one problem (IONO) on test, all other results equal.

Regarding the eGP methods without feature sampling (eGPn and eGPw), in several cases they revealed to be significantly better than their feature sampling counterparts (eGP-N and eGP-W), more often on training but also on two test cases, on problems IONO and PARKS. Even when compared to the 500 individual counterparts, they were often better on training and never worse on test. The weighted voting did not improve or worsen the obtained accuracy.

When comparing the eGP methods with the M3GP baseline, we realize that on training accuracy M3GP is better than all eGP methods on four problems (GAMETES, HEART, PARKS and SONAR), worse than all eGP methods on two problems (BCW and PPI), and on the remaining problems it is better or equal to most eGP methods, except one case where it is worse (than eGPw, on BRAZIL). On test accuracy M3GP is better than all eGP methods on four problems (IONO, PARKS, PPI and SONAR), statistically the same as all eGP methods on three problems (BCW, GAMETES, HEART), and on the remaining problem M3GP is better than all eGP feature sampling methods and statistically the same as eGPn and eGPw.

When comparing the eGP methods with the state-of-the-art RF and XG, on training both are significantly better than practically all eGP methods on all problems (except SONAR, where RF is significantly worse than all except eGP-N and eGP-W). On test accuracy, on two problems (BCW and GAMETES) there are few significant differences (XG is better than eGP-N and eGP-W), on two other problems (IONO, PARKS) both RF and XG are better than all eGP methods, and on the remaining problems RF is either the same (BRAZIL and PPI), worse (HEART) or better (SONAR) in most cases, while XG is better in all except a few cases (eGPn and eGPw on BRAZIL, eGP-N on HEART, with no significant differences).

\begin{figure}
	\centering
	\includegraphics[width=\linewidth]{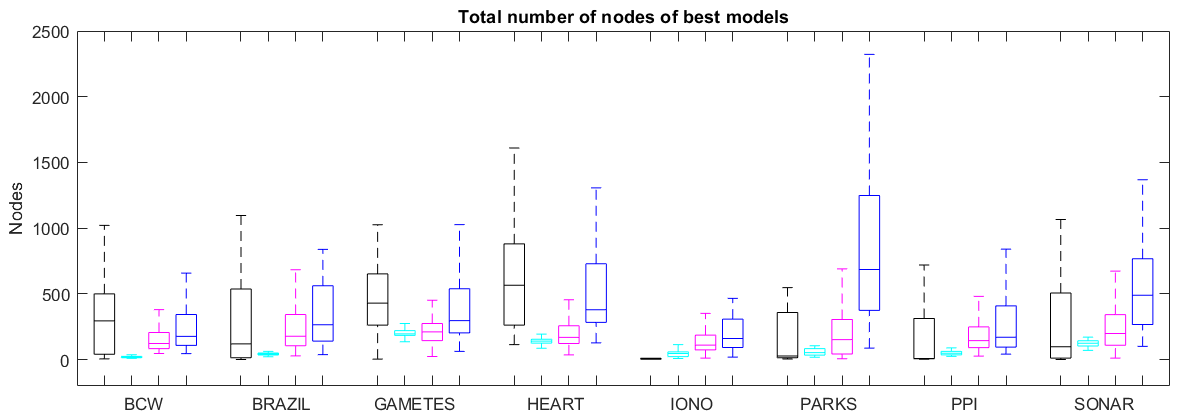}
	\vspace{-15pt}
	\caption{Number of nodes of final models. For each problem, the four boxes are: GP (black), M3GP (cyan), eGP-N $+$ eGP-W $+$ eGP-N5 $+$ eGP-W5 all together (magenta), and eGPn $+$ eGPw together (blue). All outliers removed for visualization purposes.}
	\label{nodes_boxplot}
\end{figure}

\section{Discussion}
\label{section:discussion}
In order to better understand how each of the 10 methods scored relatively to each other, we have counted how many significantly better results each one obtained among all $72 + 72 = 144$ (training $+$ test) pairwise comparisons on all problems. Table~\ref{table:countings} shows the counting (totals are the sum of all problems) and ranks the methods according to the test totals (training $+$ test in case of tie). These numbers confirm what had already been observed in the boxplots: 1) the eGP methods, although better than standard GP, were not able to outperform M3GP or the state-of-the-art RF and XG, 2) the eGP variants without feature sampling (eGPn and eGPw) are better than the other eGP methods, and 3) normal voting is generally better than weighted voting.

\begin{table}[t]
\centering
\caption{Counting of how many significantly better results each method obtained among all pairwise comparisons. The totals are the sum for all problems. Order of the problems: BCW, BRAZIL, GAMETES, HEART, IONO, PARKS, PPI, SONAR.}
\begin{tabular}{lll}
\hline
\textbf{~Method} & \multicolumn{1}{c}{\textbf{Training~~~~~~~}}        & \multicolumn{1}{c}{\textbf{Test~~~~~}}             \\ \hline
~~XG     & \hspace{3pt}3+9+9+7+9+9+8+9 = \textbf{63}\hspace{3pt}       & \hspace{3pt}2+6+0+8+8+7+8+8 = \textbf{47}~   \\
~~M3GP   & \hspace{3pt}1+1+7+8+5+7+1+8 = \textbf{38}\hspace{3pt}       & \hspace{3pt}0+6+0+1+7+7+8+8 = \textbf{37}   \\
~~RF     & \hspace{3pt}9+6+8+9+8+8+9+1 = \textbf{58}\hspace{3pt}       & \hspace{3pt}0+1+0+0+8+8+2+4 = \textbf{23}   \\
~~eGPw   & \hspace{3pt}5+5+3+5+1+4+2+4 = \textbf{29}\hspace{3pt}       & \hspace{3pt}0+1+0+1+2+1+1+0 = \textbf{~6}    \\
~~eGPn   & \hspace{3pt}2+1+5+5+5+4+3+4 = \textbf{29}\hspace{3pt}       & \hspace{3pt}0+1+0+1+2+0+1+0 = \textbf{~5}    \\
~~eGP-N5 & \hspace{3pt}2+1+2+2+5+3+3+4 = \textbf{22}\hspace{3pt}       & \hspace{3pt}0+1+0+1+2+0+1+0 = \textbf{~5}    \\
~~eGP-W5~~~~ & \hspace{3pt}4+1+1+3+2+1+2+4 = \textbf{18}\hspace{3pt}~~~~       & \hspace{3pt}0+1+0+1+2+0+1+0 = \textbf{~5}    \\
~~eGP-N  & \hspace{3pt}2+1+1+1+2+1+2+1 = \textbf{11}\hspace{3pt}       & \hspace{3pt}0+1+0+1+1+0+1+0 = \textbf{~4}    \\
~~eGP-W  & \hspace{3pt}2+1+1+1+1+0+2+1 = \textbf{~9}\hspace{3pt}       & \hspace{3pt}0+1+0+0+0+0+1+0 = \textbf{~2}    \\
~~GP     & \hspace{3pt}0+0+0+0+0+0+0+0 = \textbf{~0}\hspace{3pt}       & \hspace{3pt}0+0+0+1+0+0+0+0 = \textbf{~1}   \\
\hline
\end{tabular}
\label{table:countings}
\end{table}

Not being an ensemble method, it is noteworthy how well M3GP scored, better than RF and all other methods except XG. It is also important to emphasize that the only methods where the running parameters were tuned by cross-validation were RF and XG (see Section~\ref{subsection:parameters}). Therefore, we have no doubt regarding the superiority of M3GP over RF, and raise the question of whether it could surpass XG had its parameters also been tuned. 

Regarding the ranking of the eGP methods, it is possible that feature sampling is not necessary for a GP ensemble, due to the feature selection that most GP trees naturally do. We must also consider that the feature replacement performed by eCrossover may have highly destructive effects on the fitness of the offspring. Another thing to consider is the possible inadequacy of our certainty measure to weight the voting of the ensembles.

Although the results of the eGP methods seem disappointing, they are no doubt a viable alternative to standard GP, not only in terms of fitness but also in terms of the size of the evolved models. Figure~\ref{nodes_boxplot} shows the total number of nodes of the best models found by the GP-based methods, grouped in four sets: 1) GP only (black); 2) M3GP only (cyan); 3) eGP methods with feature sampling (eGP-N$+$eGP-W$+$eGP-N5$+$eGP-W5 all together, magenta); 4) eGP methods without feature sampling (eGPn$+$eGPw together, blue), results per problem.

The variants with feature sampling exhibit values with much less dispersion than the ones produced by GP (except on the IONO problem), and significantly lower on three problems (BCW, GAMETES, HEART). This result becomes even more important when we recall that GP used Double Tournament for bloat control (see Section~\ref{subsection:parameters}) and is composed of a single tree, while eGP did not use any bloat control and is composed of an ensemble of trees. M3GP produced the smallest solutions of all GP-based methods, however it also used Double Tournament. Regarding the number of trees that form the evolved ensembles (not shown), the eGP methods revealed a remarkable consistency among the different problems, with different runs always using between 2($\pm1$) and 13($\pm2$) trees on the best forest. This is in sharp contrast to the number of dimensions used by M3GP, with some problems using as few as 1-4 (BCW) and others using as many as 11-24 (SONAR), 13-30 (HEART) and 20-36 (GAMETES).

The GAMETES problem posed the largest difficulties to all the methods, but special attention must be given to the results obtained by some of the eGP methods, precisely the ones that scored worse in general: eGP-N, eGP-W, eGP-N5, eGP-W5. Looking back at Figure~\ref{GAMETES_with_outliers}, we observe a large amount of outliers of much higher accuracy than normal. On the test data, these are by far the best results achieved, similar to the ones reported in~\cite{lacava}, and only the four mentioned eGP methods were able to achieve them. Although out of the scope of this paper, these methods were indeed the only ones able to find, among the 1000 features of this problem, the right combinations that allowed such a big ``jump'' in accuracy. Therefore, they deserve more investigation, despite their apparent modest performance.

\section{Conclusions and Future Work}
\label{section:conclusions}

We have developed a new GP method called Ensemble GP (eGP) and tested it on eight binary classification problems from various domains, with a different number of features and observations. Different variants of eGP were compared to standard GP and M3GP baselines, and to the Random Forests and XGBoost state-of-the-art methods. The results show that eGP consistently evolves smaller and more accurate models than standard GP. M3GP and XGBoost were the best methods overall, but on a particularly hard problem the eGP variants were able to reach exceptionally good generalization results, way above all the other methods. 

As future work, we will investigate ways to improve eGP in different fronts, making it more competitive with M3GP and XGBoost while maintaining the characteristics that granted its current success. For example, bloat control and some parameter tuning are two elements that other methods are benefiting from, and that we will incorporate also in eGP. Different voting schemes may also prove beneficial, as well as alternative ways to sample features and observations. Additionally, we will also work towards extending eGP in order to give it the ability to address also regression problems and multiclass classification problems.

\section*{Acknowledgement}
This work was partially supported by FCT through funding of LASIGE
Research Unit UIDB/00408/2020 and projects
PTDC/CCI-INF/29168/2017, PTDC/CCI-CIF/29877/2017,
DSAIPA/DS/0022/2018, PTDC/ASP-PLA/28726/2017 and
PTDC/CTA-AMB/30056/2017.

\bibliographystyle{splncs04}
\bibliography{bibliografia}

\end{document}


\appendix

\begin{table}[]
\caption{BCW p-values}
\resizebox{\textwidth}{!}{\begin{tabular}{r | cccccccccccccc r }
\bf BCW&\hfil\small{GP}&\small{M3GP}&\small{eGP-N}&\small{eGP-W}&\small{eGP-N5}&\small{eGP-W5}&\small{eGPn}&\small{eGPw}&\small{RF}&\small{XG}&\\
\cmidrule{1-11}
\small{GP}&-&\textbf{0.000}&\textbf{0.000}&\textbf{0.000}&\textbf{0.000}&\textbf{0.000}&\textbf{0.000}&\textbf{0.000}&\textbf{0.000}&\multicolumn{1}{c|}{\textbf{0.000}}&\\
\small{M3GP}&0.689&-&\textbf{0.000}&\textbf{0.000}&\textbf{0.000}&\textbf{0.000}&\textbf{0.000}&\textbf{0.000}&\textbf{0.000}&\multicolumn{1}{c|}{\textbf{0.000}}&\\
\small{eGP-N}&0.484&0.224&-&0.553&0.176&0.013&0.894&\textbf{0.001}&\textbf{0.000}&\multicolumn{1}{c|}{0.088}&\\
\small{eGP-W}&0.224&0.070&0.594&-&0.012&\textbf{0.003}&0.953&\textbf{0.000}&\textbf{0.000}&\multicolumn{1}{c|}{0.029}&\\
\small{eGP-N5}&0.571&0.335&0.698&0.441&-&0.110&0.074&0.015&\textbf{0.000}&\multicolumn{1}{c|}{0.699}&\\
\small{eGP-W5}&0.324&0.870&0.076&0.075&0.086&-&\underline{\textbf{0.006}}&0.225&\textbf{0.000}&\multicolumn{1}{c|}{0.766}&\\
\small{eGPn}&0.859&0.567&0.441&0.135&0.505&0.976&-&\textbf{0.000}&\textbf{0.000}&\multicolumn{1}{c|}{\textbf{0.002}}&\rot{\rlap{\footnotesize{\ T\ R\ A\ I\ N}}}\\
\small{eGPw}&0.829&0.894&0.254&0.138&0.483&0.481&0.836&-&\textbf{0.000}&\multicolumn{1}{c|}{0.057}&\\
\small{RF}&0.812&0.677&0.122&0.014&0.357&0.812&0.563&0.688&-&\multicolumn{1}{c|}{\underline{\textbf{0.000}}}&\\
\small{XG}&0.104&0.177&\underline{\textbf{0.008}}&\underline{\textbf{0.002}}&0.014&0.449&0.192&0.111&0.046&\multicolumn{1}{c|}{-}&\\
\cmidrule{2-11}
\multicolumn{1}{r}{}&\multicolumn{11}{c}{\footnotesize{T\ E\ S\ T}} & & \\
\end{tabular}}
\end{table}

\begin{table}[]
\caption{BRAZIL p-values}
\resizebox{\textwidth}{!}{\begin{tabular}{r | cccccccccccccc r }
\bf BRAZIL&\hfil\small{GP}&\small{M3GP}&\small{eGP-N}&\small{eGP-W}&\small{eGP-N5}&\small{eGP-W5}&\small{eGPn}&\small{eGPw}&\small{RF}&\small{XG}&\\
\cmidrule{1-11}
\small{GP}&-&\textbf{0.000}&\textbf{0.000}&\textbf{0.000}&\textbf{0.000}&\textbf{0.000}&\textbf{0.000}&\textbf{0.000}&\textbf{0.000}&\multicolumn{1}{c|}{\textbf{0.000}}&\\
\small{M3GP}&\underline{\textbf{0.000}}&-&0.953&0.187&0.894&0.177&0.756&\textbf{0.003}&\textbf{0.000}&\multicolumn{1}{c|}{\textbf{0.000}}&\\
\small{eGP-N}&\underline{\textbf{0.000}}&\textbf{0.000}&-&0.212&0.947&0.244&0.678&\textbf{0.007}&\textbf{0.000}&\multicolumn{1}{c|}{\textbf{0.000}}&\\
\small{eGP-W}&\underline{\textbf{0.000}}&\textbf{0.001}&0.688&-&0.179&0.029&0.088&\textbf{0.000}&\textbf{0.000}&\multicolumn{1}{c|}{\textbf{0.000}}&\\
\small{eGP-N5}&\underline{\textbf{0.001}}&\textbf{0.003}&0.224&0.317&-&0.285&0.699&\textbf{0.008}&\textbf{0.001}&\multicolumn{1}{c|}{\textbf{0.000}}&\\
\small{eGP-W5}&\underline{\textbf{0.000}}&\textbf{0.001}&0.666&0.864&0.427&-&0.472&0.126&0.011&\multicolumn{1}{c|}{\textbf{0.000}}&\\
\small{eGPn}&\underline{\textbf{0.000}}&0.665&0.047&0.056&0.033&0.030&-&0.027&\textbf{0.003}&\multicolumn{1}{c|}{\textbf{0.000}}&\rot{\rlap{\footnotesize{\ T\ R\ A\ I\ N}}}\\
\small{eGPw}&\underline{\textbf{0.000}}&0.120&0.364&0.268&0.104&0.203&0.439&-&0.293&\multicolumn{1}{c|}{\textbf{0.000}}&\\
\small{RF}&\underline{\textbf{0.000}}&\textbf{0.000}&0.389&0.906&0.292&0.882&0.019&0.279&-&\multicolumn{1}{c|}{\textbf{0.000}}&\\
\small{XG}&\underline{\textbf{0.000}}&0.633&\underline{\textbf{0.003}}&\underline{\textbf{0.003}}&\underline{\textbf{0.003}}&\underline{\textbf{0.001}}&0.291&0.045&\underline{\textbf{0.000}}&\multicolumn{1}{c|}{-}&\\
\cmidrule{2-11}
\multicolumn{1}{r}{}&\multicolumn{11}{c}{\footnotesize{T\ E\ S\ T}} & & \\
\end{tabular}}
\end{table}

\begin{table}[]
\caption{GAMETES p-values}
\resizebox{\textwidth}{!}{\begin{tabular}{r | cccccccccccccc r }
\bf GAMETES&\hfil\small{GP}&\small{M3GP}&\small{eGP-N}&\small{eGP-W}&\small{eGP-N5}&\small{eGP-W5}&\small{eGPn}&\small{eGPw}&\small{RF}&\small{XG}&\\
\cmidrule{1-11}
\small{GP}&-&\textbf{0.000}&\textbf{0.000}&\textbf{0.000}&\textbf{0.000}&\textbf{0.000}&\textbf{0.000}&\textbf{0.000}&\textbf{0.000}&\multicolumn{1}{c|}{\textbf{0.000}}&\\
\small{M3GP}&0.468&-&\underline{\textbf{0.000}}&\underline{\textbf{0.000}}&\underline{\textbf{0.000}}&\underline{\textbf{0.000}}&\underline{\textbf{0.000}}&\underline{\textbf{0.000}}&\textbf{0.000}&\multicolumn{1}{c|}{\textbf{0.000}}&\\
\small{eGP-N}&0.544&0.679&-&0.520&\textbf{0.003}&0.077&\textbf{0.000}&\textbf{0.000}&\textbf{0.000}&\multicolumn{1}{c|}{\textbf{0.000}}&\\
\small{eGP-W}&0.853&0.450&0.739&-&0.019&0.071&\textbf{0.001}&\textbf{0.006}&\textbf{0.000}&\multicolumn{1}{c|}{\textbf{0.000}}&\\
\small{eGP-N5}&0.077&0.505&0.231&0.153&-&0.092&\textbf{0.006}&0.468&\textbf{0.000}&\multicolumn{1}{c|}{\textbf{0.000}}&\\
\small{eGP-W5}&0.529&0.156&0.151&0.487&0.025&-&\textbf{0.000}&0.027&\textbf{0.000}&\multicolumn{1}{c|}{\textbf{0.000}}&\\
\small{eGPn}&0.336&0.625&0.918&0.706&0.195&0.120&-&0.038&\textbf{0.000}&\multicolumn{1}{c|}{\textbf{0.000}}&\rot{\rlap{\footnotesize{\ T\ R\ A\ I\ N}}}\\
\small{eGPw}&0.496&0.605&0.877&0.853&0.158&0.173&0.762&-&\textbf{0.000}&\multicolumn{1}{c|}{\textbf{0.000}}&\\
\small{RF}&0.340&0.164&0.117&0.487&0.025&0.631&0.122&0.105&-&\multicolumn{1}{c|}{\textbf{0.000}}&\\
\small{XG}&0.982&0.424&0.589&1.000&0.128&0.520&0.437&0.520&0.348&\multicolumn{1}{c|}{-}&\\
\cmidrule{2-11}
\multicolumn{1}{r}{}&\multicolumn{11}{c}{\footnotesize{T\ E\ S\ T}} & & \\
\end{tabular}}
\end{table}

\begin{table}[]
\caption{HEART p-values}
\resizebox{\textwidth}{!}{\begin{tabular}{r | cccccccccccccc r }
\bf HEART&\hfil\small{GP}&\small{M3GP}&\small{eGP-N}&\small{eGP-W}&\small{eGP-N5}&\small{eGP-W5}&\small{eGPn}&\small{eGPw}&\small{RF}&\small{XG}&\\
\cmidrule{1-11}
\small{GP}&-&\textbf{0.000}&\textbf{0.000}&\textbf{0.000}&\textbf{0.000}&\textbf{0.000}&\textbf{0.000}&\textbf{0.000}&\textbf{0.000}&\multicolumn{1}{c|}{\textbf{0.000}}&\\
\small{M3GP}&0.976&-&\underline{\textbf{0.000}}&\underline{\textbf{0.000}}&\underline{\textbf{0.000}}&\underline{\textbf{0.000}}&\underline{\textbf{0.000}}&\underline{\textbf{0.000}}&\textbf{0.000}&\multicolumn{1}{c|}{\underline{\textbf{0.000}}}&\\
\small{eGP-N}&0.064&0.037&-&0.645&0.029&\textbf{0.002}&\textbf{0.000}&\textbf{0.000}&\textbf{0.000}&\multicolumn{1}{c|}{\textbf{0.000}}&\\
\small{eGP-W}&0.959&0.935&0.132&-&\textbf{0.005}&\textbf{0.000}&\textbf{0.000}&\textbf{0.000}&\textbf{0.000}&\multicolumn{1}{c|}{\textbf{0.000}}&\\
\small{eGP-N5}&0.917&0.917&0.057&0.830&-&0.260&\textbf{0.000}&\textbf{0.000}&\textbf{0.000}&\multicolumn{1}{c|}{\textbf{0.000}}&\\
\small{eGP-W5}&0.594&0.836&0.164&0.678&0.544&-&\textbf{0.000}&\textbf{0.000}&\textbf{0.000}&\multicolumn{1}{c|}{\textbf{0.000}}&\\
\small{eGPn}&0.522&0.504&0.180&0.557&0.489&0.929&-&0.894&\textbf{0.000}&\multicolumn{1}{c|}{\textbf{0.000}}&\rot{\rlap{\footnotesize{\ T\ R\ A\ I\ N}}}\\
\small{eGPw}&0.343&0.390&0.260&0.407&0.374&0.300&0.744&-&\textbf{0.000}&\multicolumn{1}{c|}{\textbf{0.000}}&\\
\small{RF}&\textbf{0.002}&\textbf{0.001}&\textbf{0.000}&0.037&\textbf{0.003}&\textbf{0.005}&\textbf{0.000}&\textbf{0.000}&-&\multicolumn{1}{c|}{\underline{\textbf{0.000}}}&\\
\small{XG}&\underline{\textbf{0.001}}&\underline{\textbf{0.001}}&0.061&\underline{\textbf{0.004}}&\underline{\textbf{0.001}}&\underline{\textbf{0.001}}&\underline{\textbf{0.002}}&\underline{\textbf{0.005}}&\underline{\textbf{0.000}}&\multicolumn{1}{c|}{-}&\\
\cmidrule{2-11}
\multicolumn{1}{r}{}&\multicolumn{11}{c}{\footnotesize{T\ E\ S\ T}} & & \\
\end{tabular}}
\end{table}

\begin{table}[]
\caption{IONO p-values}
\resizebox{\textwidth}{!}{\begin{tabular}{r | cccccccccccccc r }
\bf IONO&\hfil\small{GP}&\small{M3GP}&\small{eGP-N}&\small{eGP-W}&\small{eGP-N5}&\small{eGP-W5}&\small{eGPn}&\small{eGPw}&\small{RF}&\small{XG}&\\
\cmidrule{1-11}
\small{GP}&-&\textbf{0.000}&\textbf{0.000}&\textbf{0.000}&\textbf{0.000}&\textbf{0.000}&\textbf{0.000}&\textbf{0.000}&\textbf{0.000}&\multicolumn{1}{c|}{\textbf{0.000}}&\\
\small{M3GP}&\underline{\textbf{0.000}}&-&\underline{\textbf{0.000}}&\underline{\textbf{0.000}}&0.261&\underline{\textbf{0.000}}&0.062&\underline{\textbf{0.000}}&\textbf{0.000}&\multicolumn{1}{c|}{\textbf{0.000}}&\\
\small{eGP-N}&\underline{\textbf{0.000}}&\textbf{0.000}&-&\underline{\textbf{0.010}}&\textbf{0.000}&0.959&\textbf{0.001}&0.906&\textbf{0.000}&\multicolumn{1}{c|}{\textbf{0.000}}&\\
\small{eGP-W}&0.033&\textbf{0.000}&0.018&-&\textbf{0.000}&\textbf{0.008}&\textbf{0.000}&0.012&\textbf{0.000}&\multicolumn{1}{c|}{\textbf{0.000}}&\\
\small{eGP-N5}&\underline{\textbf{0.000}}&\textbf{0.001}&0.151&\underline{\textbf{0.001}}&-&\underline{\textbf{0.000}}&0.403&\underline{\textbf{0.000}}&\textbf{0.000}&\multicolumn{1}{c|}{\textbf{0.000}}&\\
\small{eGP-W5}&\underline{\textbf{0.000}}&\textbf{0.000}&0.198&\underline{\textbf{0.000}}&0.865&-&\textbf{0.001}&0.888&\textbf{0.000}&\multicolumn{1}{c|}{\textbf{0.000}}&\\
\small{eGPn}&\underline{\textbf{0.000}}&\textbf{0.000}&0.365&\underline{\textbf{0.002}}&0.705&0.959&-&\underline{\textbf{0.002}}&\textbf{0.000}&\multicolumn{1}{c|}{\textbf{0.000}}&\rot{\rlap{\footnotesize{\ T\ R\ A\ I\ N}}}\\
\small{eGPw}&\underline{\textbf{0.000}}&\textbf{0.000}&0.423&\underline{\textbf{0.003}}&0.750&0.750&0.784&-&\textbf{0.000}&\multicolumn{1}{c|}{\textbf{0.000}}&\\
\small{RF}&\underline{\textbf{0.000}}&\underline{\textbf{0.000}}&\underline{\textbf{0.000}}&\underline{\textbf{0.000}}&\underline{\textbf{0.000}}&\underline{\textbf{0.000}}&\underline{\textbf{0.000}}&\underline{\textbf{0.000}}&-&\multicolumn{1}{c|}{\textbf{0.000}}&\\
\small{XG}&\underline{\textbf{0.000}}&\underline{\textbf{0.000}}&\underline{\textbf{0.000}}&\underline{\textbf{0.000}}&\underline{\textbf{0.000}}&\underline{\textbf{0.000}}&\underline{\textbf{0.000}}&\underline{\textbf{0.000}}&0.553&\multicolumn{1}{c|}{-}&\\
\cmidrule{2-11}
\multicolumn{1}{r}{}&\multicolumn{11}{c}{\footnotesize{T\ E\ S\ T}} & & \\
\end{tabular}}
\end{table}

\begin{table}[]
\caption{PARKS p-values}
\resizebox{\textwidth}{!}{\begin{tabular}{r | cccccccccccccc r }
\bf PARKS&\hfil\small{GP}&\small{M3GP}&\small{eGP-N}&\small{eGP-W}&\small{eGP-N5}&\small{eGP-W5}&\small{eGPn}&\small{eGPw}&\small{RF}&\small{XG}&\\
\cmidrule{1-11}
\small{GP}&-&\textbf{0.000}&0.496&0.012&\textbf{0.000}&0.070&\textbf{0.000}&\textbf{0.000}&\textbf{0.000}&\multicolumn{1}{c|}{\textbf{0.000}}&\\
\small{M3GP}&\underline{\textbf{0.000}}&-&\underline{\textbf{0.000}}&\underline{\textbf{0.000}}&\underline{\textbf{0.000}}&\underline{\textbf{0.000}}&\underline{\textbf{0.000}}&\underline{\textbf{0.000}}&\textbf{0.002}&\multicolumn{1}{c|}{\textbf{0.000}}&\\
\small{eGP-N}&0.630&\textbf{0.000}&-&\underline{\textbf{0.004}}&\textbf{0.001}&0.267&\textbf{0.000}&\textbf{0.000}&\textbf{0.000}&\multicolumn{1}{c|}{\textbf{0.000}}&\\
\small{eGP-W}&0.342&\textbf{0.000}&0.640&-&\textbf{0.000}&\textbf{0.001}&\textbf{0.000}&\textbf{0.000}&\textbf{0.000}&\multicolumn{1}{c|}{\textbf{0.000}}&\\
\small{eGP-N5}&0.929&\textbf{0.000}&0.750&0.436&-&0.046&0.045&0.017&\textbf{0.000}&\multicolumn{1}{c|}{\textbf{0.000}}&\\
\small{eGP-W5}&0.205&\textbf{0.000}&0.111&0.026&0.177&-&\textbf{0.000}&\textbf{0.000}&\textbf{0.000}&\multicolumn{1}{c|}{\textbf{0.000}}&\\
\small{eGPn}&0.133&\textbf{0.000}&0.095&0.020&0.183&0.953&-&0.457&\textbf{0.000}&\multicolumn{1}{c|}{\textbf{0.000}}&\rot{\rlap{\footnotesize{\ T\ R\ A\ I\ N}}}\\
\small{eGPw}&0.090&\textbf{0.000}&0.052&\underline{\textbf{0.007}}&0.110&0.847&0.952&-&\textbf{0.000}&\multicolumn{1}{c|}{\textbf{0.000}}&\\
\small{RF}&\underline{\textbf{0.000}}&\underline{\textbf{0.001}}&\underline{\textbf{0.000}}&\underline{\textbf{0.000}}&\underline{\textbf{0.000}}&\underline{\textbf{0.000}}&\underline{\textbf{0.000}}&\underline{\textbf{0.000}}&-&\multicolumn{1}{c|}{\textbf{0.000}}&\\
\small{XG}&\underline{\textbf{0.000}}&0.099&\underline{\textbf{0.000}}&\underline{\textbf{0.000}}&\underline{\textbf{0.000}}&\underline{\textbf{0.000}}&\underline{\textbf{0.000}}&\underline{\textbf{0.000}}&0.021&\multicolumn{1}{c|}{-}&\\
\cmidrule{2-11}
\multicolumn{1}{r}{}&\multicolumn{11}{c}{\footnotesize{T\ E\ S\ T}} & & \\
\end{tabular}}
\end{table}

\begin{table}[]
\caption{PPI p-values}
\resizebox{\textwidth}{!}{\begin{tabular}{r | cccccccccccccc r }
\bf PPI&\hfil\small{GP}&\small{M3GP}&\small{eGP-N}&\small{eGP-W}&\small{eGP-N5}&\small{eGP-W5}&\small{eGPn}&\small{eGPw}&\small{RF}&\small{XG}&\\
\cmidrule{1-11}
\small{GP}&-&\textbf{0.000}&\textbf{0.000}&\textbf{0.000}&\textbf{0.000}&\textbf{0.000}&\textbf{0.000}&\textbf{0.000}&\textbf{0.000}&\multicolumn{1}{c|}{\textbf{0.000}}&\\
\small{M3GP}&\underline{\textbf{0.000}}&-&\textbf{0.000}&\textbf{0.000}&\textbf{0.000}&\textbf{0.000}&\textbf{0.000}&\textbf{0.000}&\textbf{0.000}&\multicolumn{1}{c|}{\textbf{0.000}}&\\
\small{eGP-N}&\underline{\textbf{0.000}}&\textbf{0.000}&-&0.130&0.149&0.240&0.018&0.953&\textbf{0.000}&\multicolumn{1}{c|}{\textbf{0.000}}&\\
\small{eGP-W}&\underline{\textbf{0.000}}&\textbf{0.000}&0.128&-&\textbf{0.005}&0.017&\textbf{0.000}&0.124&\textbf{0.000}&\multicolumn{1}{c|}{\textbf{0.000}}&\\
\small{eGP-N5}&\underline{\textbf{0.000}}&\textbf{0.000}&0.912&0.249&-&0.790&0.399&0.122&\textbf{0.000}&\multicolumn{1}{c|}{\textbf{0.000}}&\\
\small{eGP-W5}&\underline{\textbf{0.000}}&\textbf{0.000}&0.301&0.375&0.574&-&0.337&0.231&\textbf{0.000}&\multicolumn{1}{c|}{\textbf{0.000}}&\\
\small{eGPn}&\underline{\textbf{0.000}}&\textbf{0.000}&0.745&0.019&0.773&0.231&-&0.014&\textbf{0.000}&\multicolumn{1}{c|}{\textbf{0.000}}&\rot{\rlap{\footnotesize{\ T\ R\ A\ I\ N}}}\\
\small{eGPw}&\underline{\textbf{0.000}}&\textbf{0.000}&0.569&0.019&0.663&0.152&0.728&-&\textbf{0.000}&\multicolumn{1}{c|}{\textbf{0.000}}&\\
\small{RF}&\underline{\textbf{0.000}}&\textbf{0.000}&0.530&\underline{\textbf{0.003}}&0.469&0.030&0.416&0.690&-&\multicolumn{1}{c|}{\underline{\textbf{0.000}}}&\\
\small{XG}&\underline{\textbf{0.000}}&0.877&\underline{\textbf{0.000}}&\underline{\textbf{0.000}}&\underline{\textbf{0.000}}&\underline{\textbf{0.000}}&\underline{\textbf{0.000}}&\underline{\textbf{0.000}}&\underline{\textbf{0.000}}&\multicolumn{1}{c|}{-}&\\
\cmidrule{2-11}
\multicolumn{1}{r}{}&\multicolumn{11}{c}{\footnotesize{T\ E\ S\ T}} & & \\
\end{tabular}}
\end{table}

\begin{table}[]
\caption{SONAR p-values}
\resizebox{\textwidth}{!}{\begin{tabular}{r | cccccccccccccc r }
\bf SONAR&\hfil\small{GP}&\small{M3GP}&\small{eGP-N}&\small{eGP-W}&\small{eGP-N5}&\small{eGP-W5}&\small{eGPn}&\small{eGPw}&\small{RF}&\small{XG}&\\
\cmidrule{1-11}
\small{GP}&-&\textbf{0.000}&\textbf{0.000}&\textbf{0.000}&\textbf{0.000}&\textbf{0.000}&\textbf{0.000}&\textbf{0.000}&\textbf{0.000}&\multicolumn{1}{c|}{\textbf{0.000}}&\\
\small{M3GP}&\underline{\textbf{0.000}}&-&\underline{\textbf{0.000}}&\underline{\textbf{0.000}}&\underline{\textbf{0.000}}&\underline{\textbf{0.000}}&\underline{\textbf{0.000}}&\underline{\textbf{0.000}}&\underline{\textbf{0.000}}&\multicolumn{1}{c|}{\textbf{0.000}}&\\
\small{eGP-N}&0.703&\textbf{0.000}&-&0.275&\textbf{0.000}&\textbf{0.000}&\textbf{0.000}&\textbf{0.000}&0.594&\multicolumn{1}{c|}{\textbf{0.000}}&\\
\small{eGP-W}&0.716&\textbf{0.000}&0.509&-&\textbf{0.000}&\textbf{0.000}&\textbf{0.000}&\textbf{0.000}&0.534&\multicolumn{1}{c|}{\textbf{0.000}}&\\
\small{eGP-N5}&0.409&\textbf{0.000}&0.295&1.000&-&0.767&0.217&0.124&\underline{\textbf{0.000}}&\multicolumn{1}{c|}{\textbf{0.000}}&\\
\small{eGP-W5}&0.432&\textbf{0.000}&0.700&0.286&0.193&-&0.321&0.225&\underline{\textbf{0.000}}&\multicolumn{1}{c|}{\textbf{0.000}}&\\
\small{eGPn}&0.019&\textbf{0.000}&0.055&0.018&0.012&0.172&-&0.603&\underline{\textbf{0.000}}&\multicolumn{1}{c|}{\textbf{0.000}}&\rot{\rlap{\footnotesize{\ T\ R\ A\ I\ N}}}\\
\small{eGPw}&0.013&\textbf{0.000}&0.034&0.012&0.011&0.110&0.760&-&\underline{\textbf{0.000}}&\multicolumn{1}{c|}{\textbf{0.000}}&\\
\small{RF}&\underline{\textbf{0.001}}&\textbf{0.000}&\underline{\textbf{0.005}}&\underline{\textbf{0.004}}&\underline{\textbf{0.002}}&0.014&0.524&0.744&-&\multicolumn{1}{c|}{\textbf{0.000}}&\\
\small{XG}&\underline{\textbf{0.000}}&0.207&\underline{\textbf{0.000}}&\underline{\textbf{0.000}}&\underline{\textbf{0.000}}&\underline{\textbf{0.000}}&\underline{\textbf{0.000}}&\underline{\textbf{0.000}}&\underline{\textbf{0.000}}&\multicolumn{1}{c|}{-}&\\
\cmidrule{2-11}
\multicolumn{1}{r}{}&\multicolumn{11}{c}{\footnotesize{T\ E\ S\ T}} & & \\
\end{tabular}}
\end{table}

\begin{figure}[]
	\centering
	\includegraphics[scale=.60]{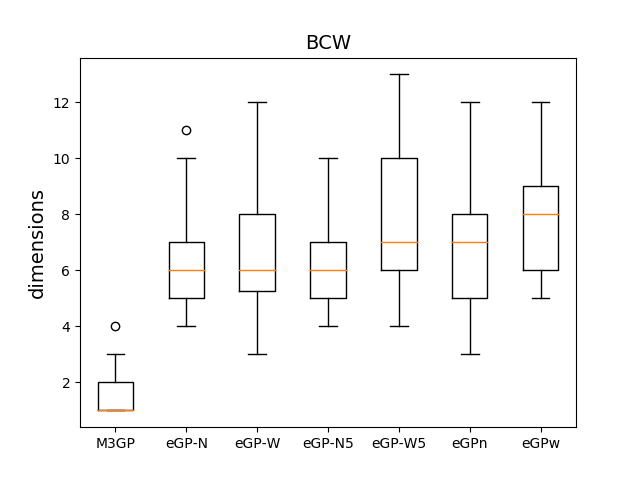}
	\caption{Boxplot for the number of dimensions with each method in the BCW dataset.}
	\label{BCW_dimensions_with_outliers}
\end{figure}

\begin{figure}[]
	\centering
	\includegraphics[scale=.60]{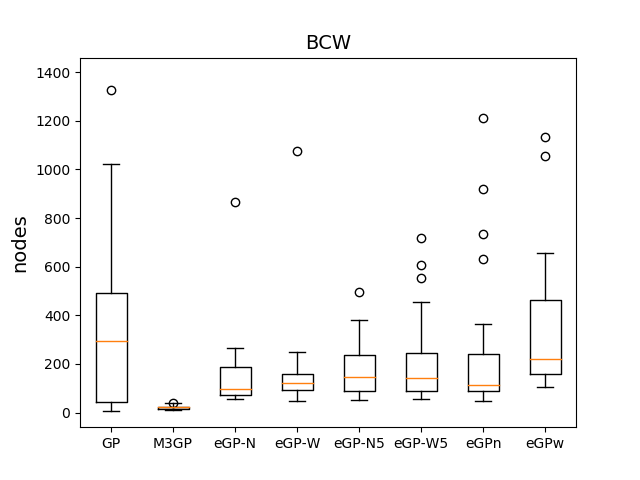}
	\caption{Boxplot for the number of nodes with each method in the BCW dataset.}
	\label{BCW_nodes_with_outliers}
\end{figure}

\begin{figure}[]
	\centering
	\includegraphics[scale=.60]{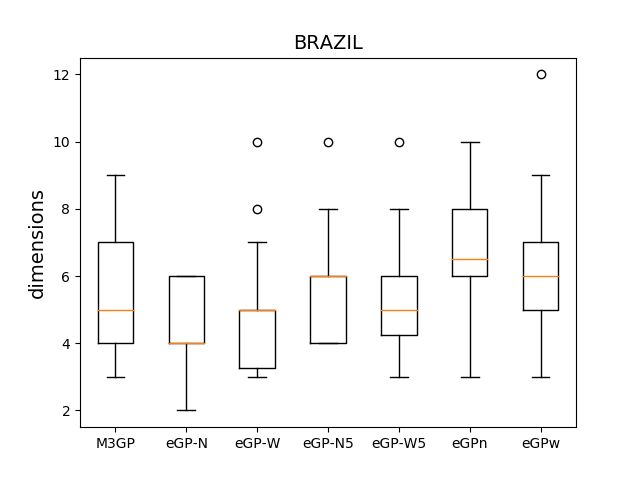}
	\caption{Boxplot for the number of dimensions with each method in the BRAZIL dataset.}
	\label{BRAZIL_dimensions_with_outliers}
\end{figure}

\begin{figure}[]
	\centering
	\includegraphics[scale=.60]{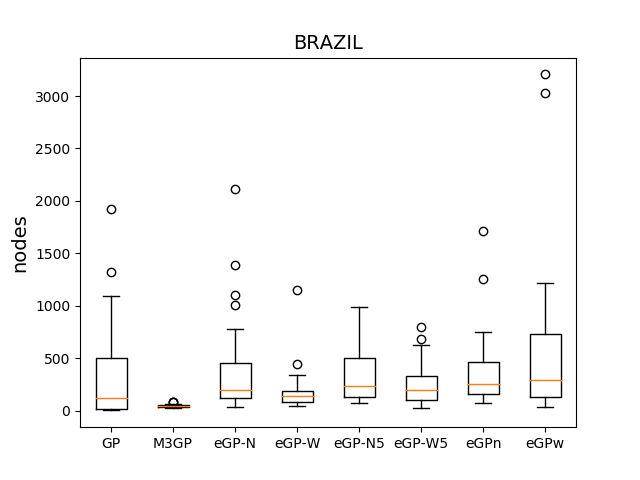}
	\caption{Boxplot for the number of nodes with each method in the BRAZIL dataset.}
	\label{BRAZIL_nodes_with_outliers}
\end{figure}

\begin{figure}[]
	\centering
	\includegraphics[scale=.60]{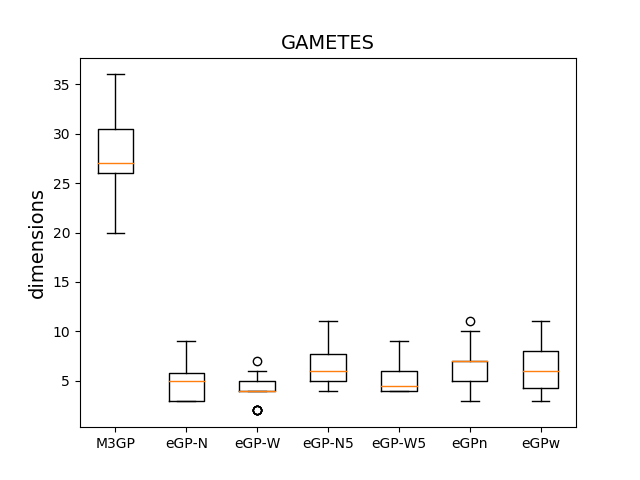}
	\caption{Boxplot for the number of dimensions with each method in the GAMETES dataset.}
	\label{GAMETES_dimensions_with_outliers}
\end{figure}

\begin{figure}[]
	\centering
	\includegraphics[scale=.60]{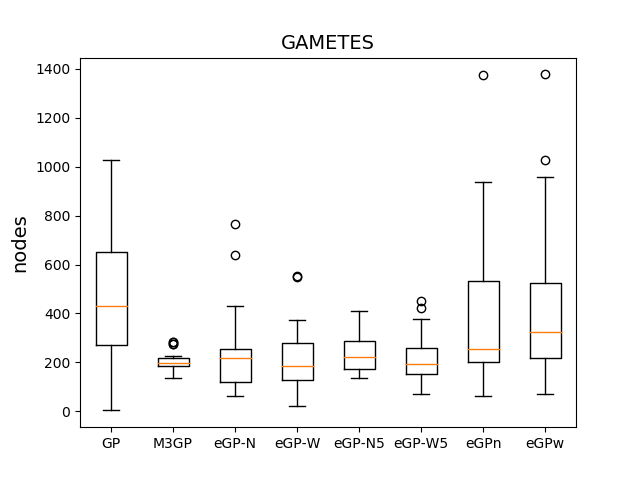}
	\caption{Boxplot for the number of nodes with each method in the GAMETES dataset.}
	\label{GAMETES_nodes_with_outliers}
\end{figure}

\begin{figure}[]
	\centering
	\includegraphics[scale=.60]{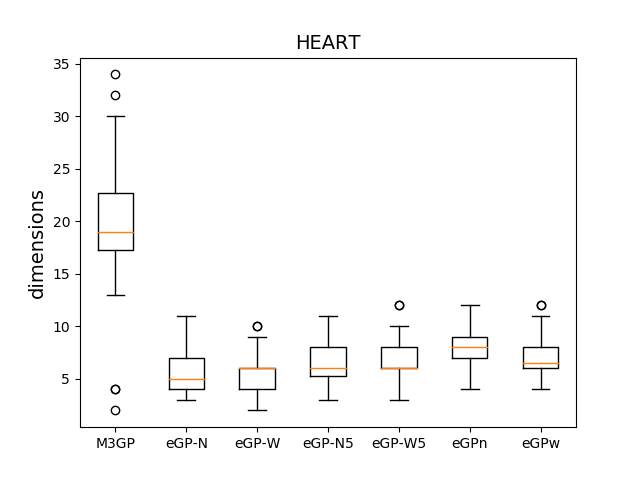}
	\caption{Boxplot for the number of dimensions with each method in the HEART dataset.}
	\label{HEART_dimensions_with_outliers}
\end{figure}

\begin{figure}[]
	\centering
	\includegraphics[scale=.60]{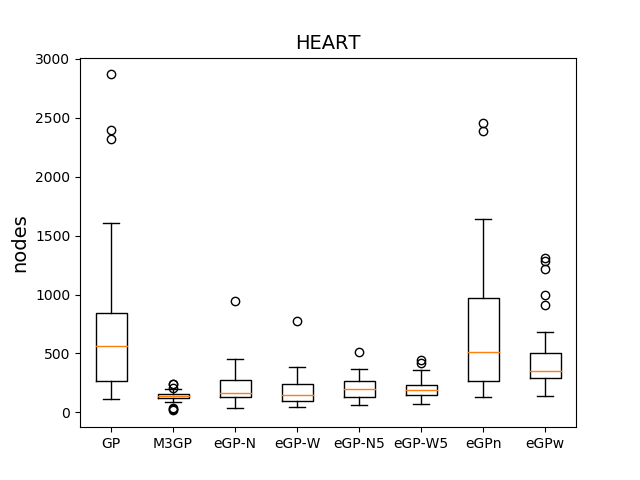}
	\caption{Boxplot for the number of nodes with each method in the HEART dataset.}
	\label{HEART_nodes_with_outliers}
\end{figure}

\begin{figure}[]
	\centering
	\includegraphics[scale=.60]{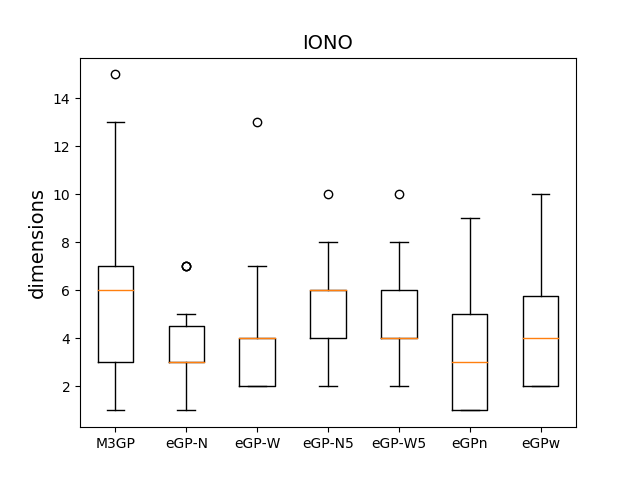}
	\caption{Boxplot for the number of dimensions with each method in the IONO dataset.}
	\label{IONO_dimensions_with_outliers}
\end{figure}

\begin{figure}[]
	\centering
	\includegraphics[scale=.60]{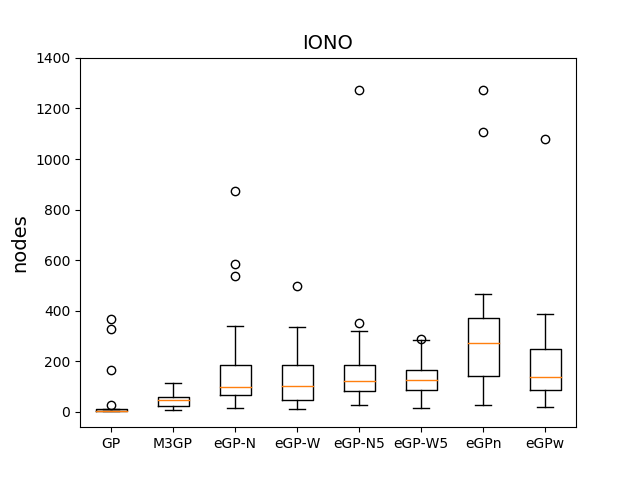}
	\caption{Boxplot for the number of nodes with each method in the IONO dataset.}
	\label{IONO_nodes_with_outliers}
\end{figure}

\begin{figure}[]
	\centering
	\includegraphics[scale=.60]{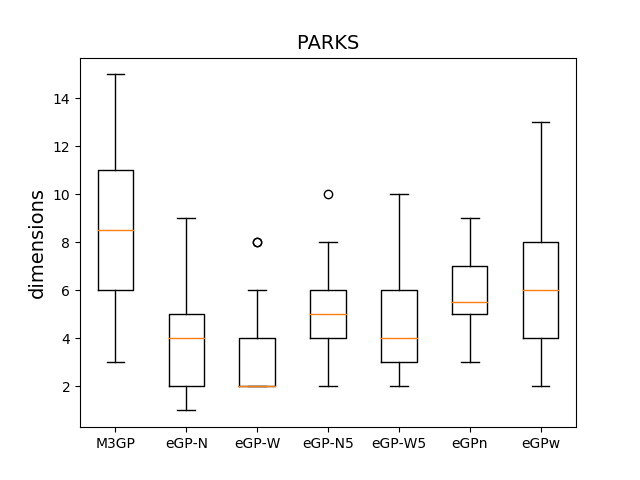}
	\caption{Boxplot for the number of dimensions with each method in the PARKS dataset.}
	\label{PARK_dimensions_with_outliers}
\end{figure}

\begin{figure}[]
	\centering
	\includegraphics[scale=.60]{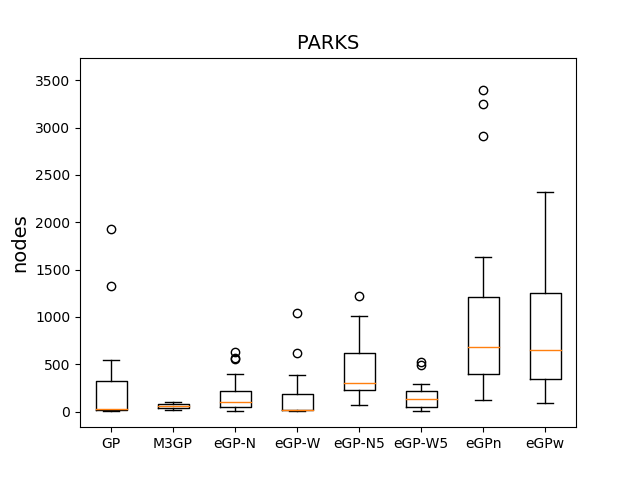}
	\caption{Boxplot for the number of nodes with each method in the PARKS dataset.}
	\label{PARK_nodes_with_outliers}
\end{figure}

\begin{figure}[]
	\centering
	\includegraphics[scale=.60]{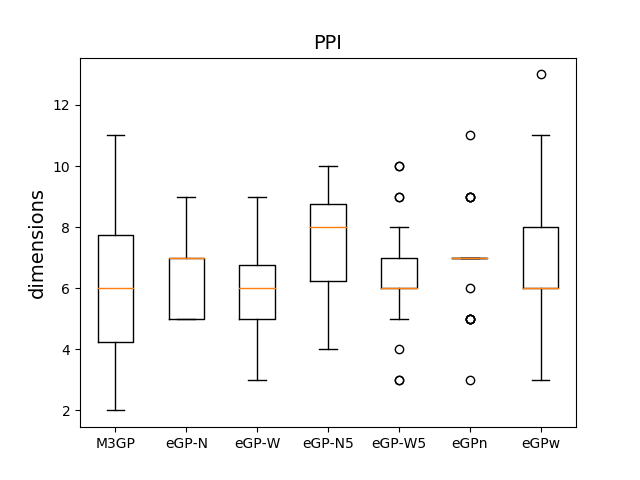}
	\caption{Boxplot for the number of dimensions with each method in the PPI dataset.}
	\label{PPI_dimensions_with_outliers}
\end{figure}

\begin{figure}[]
	\centering
	\includegraphics[scale=.60]{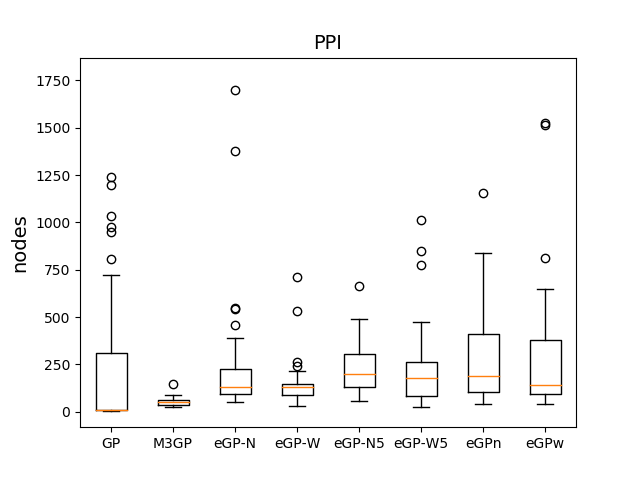}
	\caption{Boxplot for the number of nodes with each method in the PPI dataset.}
	\label{PPI_nodes_with_outliers}
\end{figure}

\begin{figure}[]
	\centering
	\includegraphics[scale=.60]{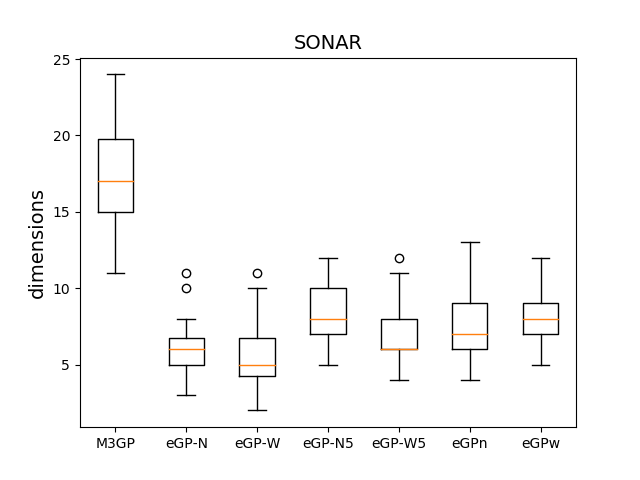}
	\caption{Boxplot for the number of dimensions with each method in the SONAR dataset.}
	\label{SONAR_dimensions_with_outliers}
\end{figure}

\begin{figure}[]
	\centering
	\includegraphics[scale=.60]{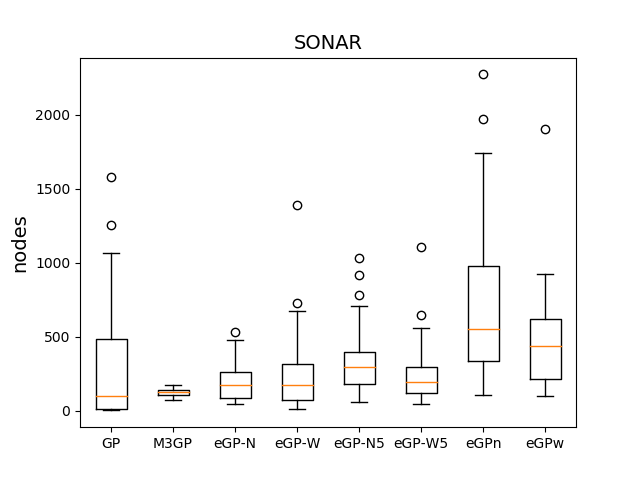}
	\caption{Boxplot for the number of nodes with each method in the SONAR dataset.}
	\label{SONAR_nodes_with_outliers}
\end{figure}